%% file: main.tex
\documentclass[runningheads]{llncs}
\usepackage{graphicx}

\usepackage{tikz}
\usepackage{comment}
\usepackage{amsmath,amssymb} 
\usepackage{color}

 \newcommand\figcaption{\def\@captype{figure}\caption}
 \newcommand\tabcaption{\def\@captype{table}\caption}
 \usepackage{subcaption}
\usepackage[accsupp]{axessibility}  

\usepackage[width=122mm,left=12mm,paperwidth=148mm,height=195mm,top=12mm,paperheight=217mm]{geometry} 

\usepackage{graphicx}
\usepackage{amsmath}
\usepackage{amssymb}
\usepackage{booktabs}
\usepackage{textcomp}  
\usepackage{amsfonts}
\usepackage{multirow}
\usepackage{algorithm}
\usepackage{algorithmic}
\usepackage{float}
\usepackage{tabularx}
\usepackage{array}
\usepackage[misc]{ifsym}
\usepackage[pagebackref=true,breaklinks=true,letterpaper=true,colorlinks,bookmarks=false]{hyperref}

\usepackage{subcaption}
\input{math_commands.tex}

\usepackage{lipsum}

\usepackage{xspace}

\newcommand*{\eg}{\emph{e.g.}}
\newcommand*{\ie}{\emph{i.e.}}

\begin{document}
\pagestyle{headings}
\mainmatter

\def\PAPERTITLE{InvPT: Inverted Pyramid Multi-task Transformer for Dense Scene Understanding}
\title{\PAPERTITLE} 

\titlerunning{Inverted Pyramid Multi-task Transformer for Dense Scene Understanding}
%

\author{
Hanrong Ye and Dan Xu\textsuperscript{\Letter}
}
\authorrunning{H. Ye and D. Xu}
%
\institute{Department of Computer Science and Engineering, HKUST\\
Clear Water Bay, Kowloon, Hong Kong\\
\email{\{hyeae,danxu\}@cse.ust.hk}
}
\maketitle
\vspace{-15PT}
\begin{abstract}
Multi-task dense scene understanding is a thriving research domain that requires simultaneous perception and reasoning on a series of correlated tasks with pixel-wise prediction. Most existing works encounter a severe limitation of modeling in the locality due to heavy utilization of convolution operations, while learning interactions and inference in a global spatial-position and multi-task context is critical for this problem.
In this paper, we propose a novel end-to-end Inverted Pyramid multi-task Transformer (\textbf{InvPT}) to perform simultaneous modeling of spatial positions and multiple tasks in a unified framework. To the best of our knowledge, this is the first work that explores designing a transformer structure for multi-task dense prediction for scene understanding. Besides, it is widely demonstrated that a higher spatial resolution is remarkably beneficial for dense predictions, while it is very challenging for existing transformers to go deeper with higher resolutions due to huge complexity to large spatial size.
InvPT presents an efficient UP-Transformer block to learn multi-task feature interaction at gradually increased resolutions, which also incorporates effective self-attention message passing and multi-scale feature aggregation to produce task-specific prediction at a high resolution.
Our method achieves superior multi-task performance on NYUD-v2 and PASCAL-Context datasets respectively, and significantly outperforms previous state-of-the-arts. The code is available at~\url{https://github.com/prismformore/InvPT}.
\end{abstract}

\section{Introduction}
Multi-task visual scene understanding typically requires joint learning and reasoning on a bunch of correlated tasks~\cite{mtlsurvey}, which is highly important in computer vision and has a wide range of application scenarios such as autonomous driving, robotics, and augmented or virtual reality (AR/VR). Many of visual scene understanding tasks produce pixel-wise predictions for dense understanding of the scene~\cite{kendall2018multi} such as semantic segmentation, monocular depth estimation, and human parsing. These dense prediction tasks essentially have abundant explicit and implicit correlation at the pixel level~\cite{padnet}, which is very beneficial and can be fully utilized to improve the overall performance of multi-task models. However, how to effectively learn and exploit the cross-task correlation (\eg~complementarity and consistency) in a single model remains a challenging open issue.

\begin{figure}[!t]
    \centering
    \includegraphics[width=0.7\linewidth]{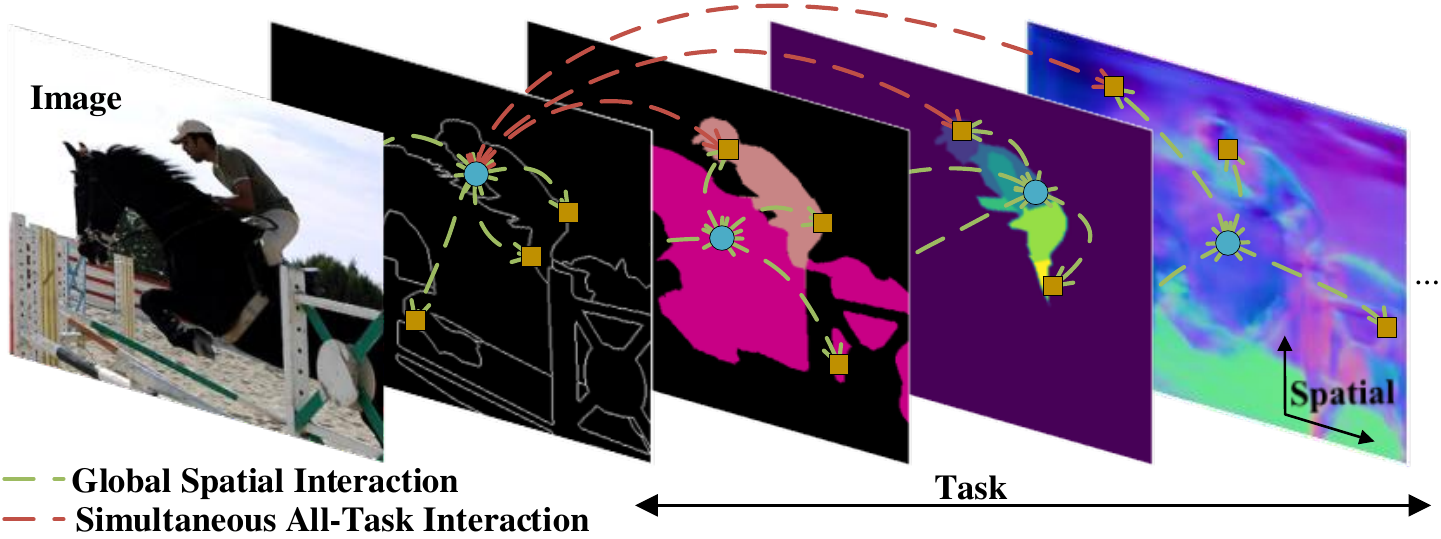}
    \vspace{-8pt}
    \caption{Joint learning and inference of global spatial interaction and simultaneous all-task interaction is critically important for multi-task dense prediction.}
    \vspace{-22pt}
    \label{fig:multi-task contexts}
\end{figure}

\par To advance the multi-task dense scene understanding, existing works mostly rely on the powerful Convolutional Neural Networks (CNN), and significant effort has been made by developing multi-task optimization losses~\cite{kendall2018multi} and designing~\cite{padnet} or searching~\cite{gao2020mtl} multi-task information sharing strategies and network structures. Although promising performance has been achieved, these works are still limited by the nature of convolution kernels that are heavily used in their deep learning frameworks, which model critical spatial and task related contexts in relatively local perceptive fields (\ie~locality discussed by previous works~\cite{NonLocal2018,attn_aug}).
Although the recently proposed attention based methods address this issue~\cite{papnet,psd,atrc}, the scope of their cross-task interaction is still highly limited. However, for multi-task dense scene understanding, the capability of capturing long-range dependency and \emph{simultaneously} modeling global relationships of \emph{all} tasks is crucially important for this pixel-wise multi-task problem (see Fig.~\ref{fig:multi-task contexts}).

\par On the other hand, recently the transformer models have been introduced to model the long-range spatial relationship for dense prediction problems~\cite{dpt,hrformer} but they only target the setting of \emph{single-task} learning, while the joint modeling of multiple dense prediction tasks with transformer is rarely explored in the literature, and it is not a trivial problem to globally model both the spatial and the cross-task correlations in a unified transformer framework. Besides, the performance of dense prediction tasks is greatly affected by the resolution of the final feature maps produced from the model, while it is very difficult for existing transformers to go deeper with higher resolution because of huge complexity brought by large spatial size, and typically many transformers downsample the spatial resolution dramatically to reduce the computation overhead~\cite{vit,pvt}.

\par To tackle the above-mentioned issues, in this work we propose a novel end-to-end Inverted Pyramid Multi-task Transformer (InvPT) framework, which can jointly model the long-range dependency within spatial and all-task contexts, and also efficiently learn fine-grained dense prediction maps at a higher resolution, for multi-task dense scene understanding. Specifically, it consists of three core designs: (i) an InvPT transformer encoder to learn generic image representations for input images, (ii) an InvPT Transformer decoder built by consecutively stacking a proposed efficient UP-Transformer block, to model implicit correlations among all the dense prediction tasks and produce multi-task features with gradually increased resolutions, and (iii) two effective multi-scale strategies, \ie~cross-scale Attention Message Passing (AMP) that aggregates self-attention maps across different transformer blocks, and multi-scale Encoder Feature Aggregation (EFA) that {enhances the decoding features with multi-scale information from the InvPT transformer encoder}.
\par The proposed method yields strong multi-task performance measured by the relative improvement metric~\cite{astmt}, which is {$2.59\%$ and $1.76\%$} on NYUD-v2 and PASCAL-Context datasets respectively.~The proposed method also \emph{largely} outperforms other state-of-the-art methods on all 9 evaluation metrics of these two benchmarks. Notably on NYUD-v2, it surpasses the best competitor by \textbf{7.23} {points}~(mIoU) on semantic segmentation, while on PASCAL-Context it outperforms the previous best result by \textbf{11.36} and \textbf{4.68} points~(mIoU) on semantic segmentation and human parsing, respectively.

\par In summary, our main contribution is three-fold:
\begin{itemize}
\vspace{-5pt}
\setlength{\parskip}{-1pt}
  \item We propose a novel end-to-end Inverted Pyramid Multi-task Transformer (InvPT) framework for jointly learning multiple dense prediction tasks, which can effectively model long-range interaction in both spatial and all-task contexts in a unified architecture. As far as we know, it is the \emph{first} work to present a transformer structure for this problem.
  \item We design an efficient UP-Transformer block, which allows for multi-task feature interaction and refinement at gradually \emph{increased} resolutions, and can construct a multi-layer InvPT decoder by consecutively stacking multiple blocks to produce final feature maps with a high resolution to largely boost dense predictions. The UP-Transformer block can also flexibly embed multi-scale information through two designed strategies, \ie~cross-scale self-attention message passing and InvPT encoder feature aggregation.
    \item The proposed framework obtains superior performance on multi-task dense prediction and \emph{remarkably} outperforms the previous state-of-the-arts on two challenging benchmarks (\ie~Pascal-Context and NYUD-v2).
\end{itemize}

\vspace{-8pt}
\section{Related Work}
\label{sec:relatedwork}
We review the most related works in the literature from two angles, \ie~multi-task deep learning for scene understanding and visual transformers.

\par\noindent\textbf{Multi-task Deep Learning for Scene Understanding} As an active research field~\cite{liang2019multi,astmt,padnet},
multi-task deep learning can greatly help to improve the efficiency of training, as it only needs to optimize once for multiple tasks, and the overall performance of scene understanding when compared with performing several scene understanding tasks separately~\cite{mti}.
Multi-task deep learning methods mainly focus on two directions~\cite{mtlsurvey}, \ie~multi-task optimization and network architecture design.
Previous works in the former direction typically investigate loss balancing techniques in the optimization process to address the problem of task competition~\cite{gradnorm,gradientsign,kendall2018multi}.
In the latter direction researchers design explicit or implicit mechanisms for modeling cross-task interaction and embed them into the whole deep model~\cite{crossstitch,sluice,nddr} or only the decoder stage~\cite{papnet,psd,zhang2021transfer}.

\par Regarding the multi-task dense scene understanding where all the tasks require pixel-wise predictions, many pioneering research works~\cite{kendall2018multi,astmt,crossstitch,nddr,padnet} have explored this field. Specifically, Xu~\textit{et al.}~\cite{padnet} propose PAD-Net with an effective information distillation module with attention guided cross-task message passing on multi-task predictions. MTI-Net~\cite{mti} designs a sophisticated  multi-scale and multi-task CNN architecture to distill information at multiple feature scales.
However, most of these methods adopt CNN to learn multi-task representation and model it in a limited local context.
To address this issue, there are also some exciting works developing attention-based mechanisms. For instance, \cite{papnet}, \cite{psd} and \cite{zhang2021transfer}
design spatial global or local attention learning strategies within each task-specific branch, and propagate the attention to refine features among all the tasks.
Concurrent to our work, \cite{atrc} proposes to build up decoder via searching for proper cross-task attention structures with neural architecture search (NAS). Despite their innovative designs, these works still fail to jointly model spatial and cross-task contexts in a global manner.

\par Different from these works, we propose a novel multi-task transformer framework to learn long-range interactions of spatial and cross-task relationships in global contexts, which is critical for this complex multi-task problem.
\begin{figure*}[!t]
	\centering
	\includegraphics[width=1\textwidth]{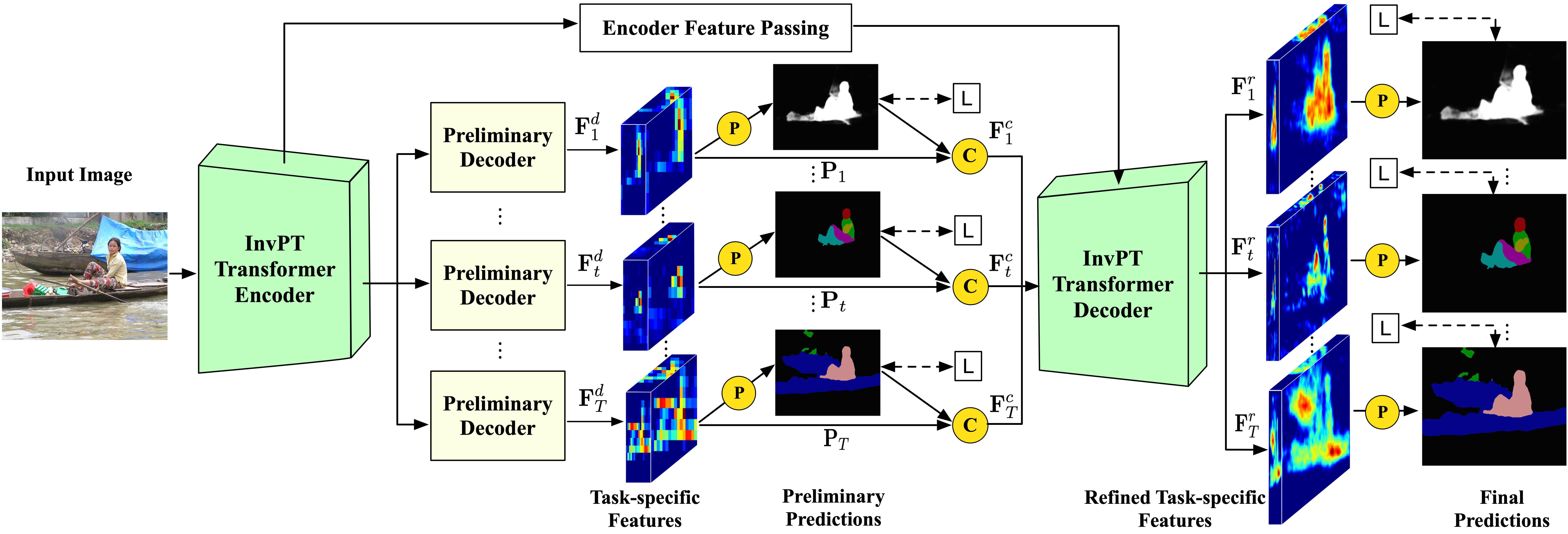}
\vspace{-18pt}
\caption{Framework overview of the proposed Inverted Pyramid Multi-task Transformer (InvPT) for dense scene understanding.
The task-shared transformer encoder learns generic visual representations from the input image. Then for each task $t\in\{ 1\dots T\}$, the preliminary decoder produces task-specific feature $\mathbf{F}^d_t$ and preliminary prediction $\mathbf{P}_t$, which are combined as $\mathbf{F}^c_t$, serving as the input of the InvPT decoder to generate \emph{refined and resolution-enlarged} task-specific features via globally modeling spatial and all-task interactions for the final prediction. \textcircled{c},  \textcircled{p} and L denote the channel-wise concatenation, linear projection layer and loss function, respectively.}
\vspace{-22pt}
\label{fig:framework}
\end{figure*}

\par\noindent\textbf{Visual Transformer}
Research interest in visual transformers grows rapidly nowadays inspired by the recent success of transformers~\cite{transformer} in multi-modal learning~\cite{clip,dalle}, 2D~\cite{parmar2018image,vit,detr,ipt,han2021transformer,chen2021pix2seq} and 3D computer vision tasks~\cite{pan20213d,mao2021voxel,Zhao_2021_ICCV}.
The transformer models are originally designed for natural language processing tasks, and then show strong performance and generalization ability in solving vision problems~\cite{vit,bhojanapalli2021understanding,bai2021transformers}. Exciting results have been achieved from different aspects including:
(i) Enhancing the self-attention mechanism to incorporate useful inductive bias~\cite{convit,ramachandran2019stand,swin,focal}. \eg~Swin Transformer~\cite{swin} replaces the global attention with shifted window attention to improve efficiency; Focal Transformer~\cite{focal} combines coarse-granularity global attention with fine-grained local attention to balance model efficiency and effectiveness;
(ii) Combining transformer with CNNs~\cite{srinivas2021bottleneck,pvt,cvt}. \eg~BOTNet~\cite{srinivas2021bottleneck} uses a specially designed multi-head self-attention head to replace final three bottleneck blocks of ResNet; PVT~\cite{pvt} and CVT~\cite{cvt} embed convolutional layers in a hierarchical transformer framework and demonstrate that they can help improve the performance;
(iii) Designing special training techniques~\cite{deit,liu2021efficient,touvron2021going}. \eg~DEIT~\cite{deit} proposes a special transformer token for knowledge distillation from CNNs; DRLOC~\cite{liu2021efficient} designs a self-supervised auxiliary task to make transformer learn spatial relations. Regarding dense scene understanding tasks~\cite{pvt,xie2021segformer}, Ranftl~\textit{et al.}~\cite{dpt} recently propose a visual transformer framework with transpose convolution for dealing with dense prediction, while HRFormer~\cite{hrformer} adopts local attention to keep multi-scale features efficiently in the network.
\par To the best of our knowledge, this is the first exploration of simultaneously modeling multiple dense prediction tasks in a unified transformer framework for scene understanding. The proposed framework jointly learns spatial and all-task interactions in a global context at gradually \emph{increased} resolutions with a novel and efficient UP-Transformer block, which can produce spatially higher-resolution feature maps to significantly boost dense predictions, and effectively incorporate multi-scale information via the proposed strategies of self-attention message passing and encoder feature aggregation.

\section{InvPT for Multi-Task Dense Prediction}
\subsection{Framework Overview}
The overall framework of the proposed Inverted Pyramid Multi-task Transformer (InvPT) is depicted in~Fig.~\ref{fig:framework}.
It mainly consists of three core parts, \ie~a task-shared InvPT transformer encoder, the task-specific preliminary decoders, and the InvPT transformer decoder. Specifically, the transformer encoder learns \emph{generic} visual representations from the input images for all tasks.~Then, the preliminary decoders produce \emph{task-specific} features and preliminary predictions, which are supervised by the ground-truth labels.
The task-specific feature and preliminary prediction of each task are combined and concatenated as a sequence serving as input of the InvPT Transformer decoder, to learn to produce refined task-specific representations within global spatial and task contexts, which are further used to produce the final predictions with task-specific linear projection layers. The details of these parts are introduced as follows.

\subsection{InvPT Transformer Encoder}
The transformer encoder is shared for different tasks to learn generic visual representations from the input image. The self-attention mechanism of the transformer can help to learn a global feature representation of the input via long-range modeling of the spatial dependency of image pixels or patches.
In our implementation, we consider different alternatives for the encoder including ViT~\cite{vit} and Swin Transformer~\cite{swin}. We obtain from the encoder a feature sequence and reshape it as a spatial feature map with resolution $H_0\times W_0$ where $H_0$ and $W_0$ denote its height and width respectively. The feature map is then input into $T$ preliminary decoders to learn the $T$ task-specific feature maps.

\begin{figure}[t]
\vspace{-10pt}
\centering
\includegraphics[width=0.85\textwidth]{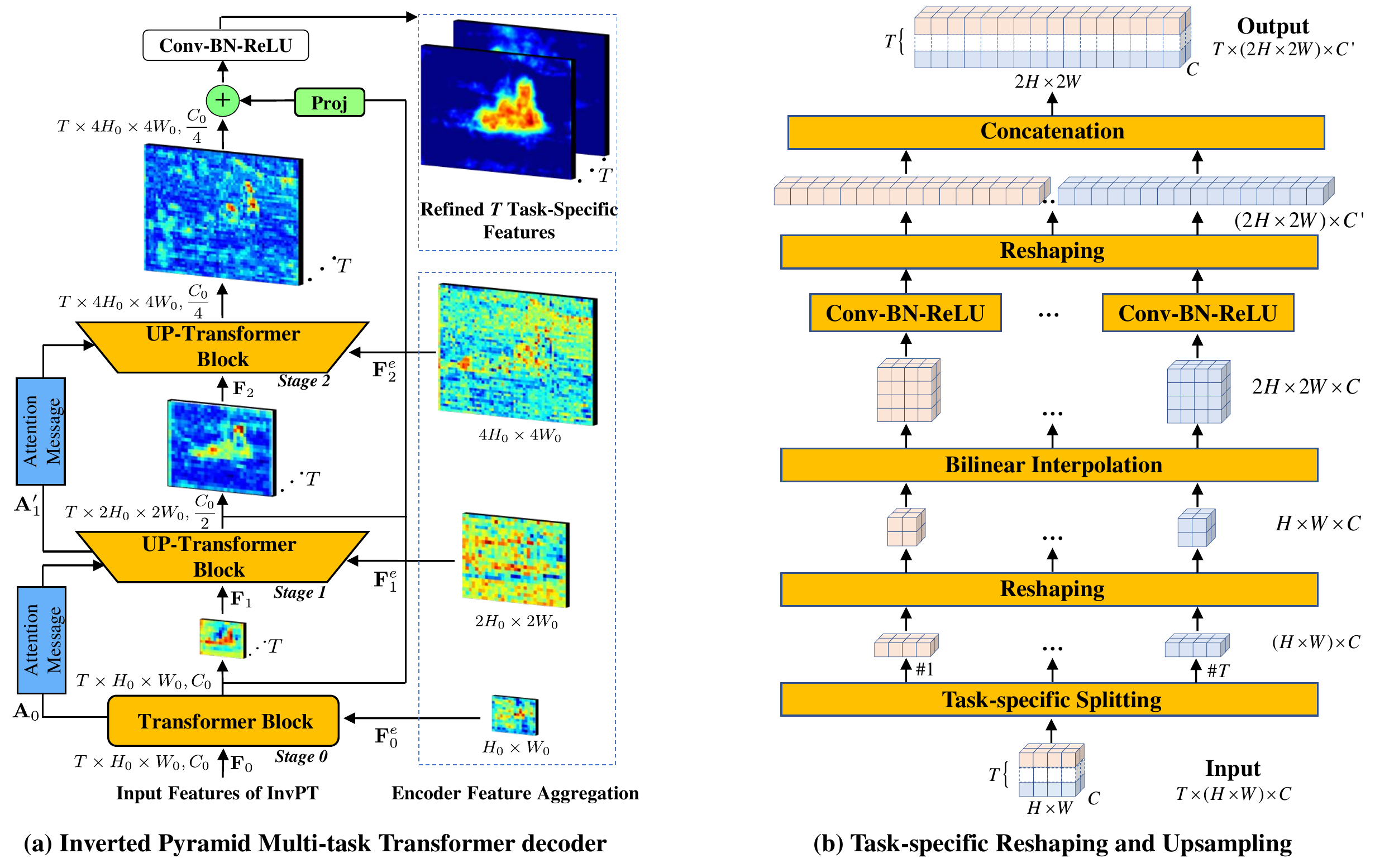}
\vspace{-10pt}
\caption{\textbf{(a)} Illustration of InvPT Transformer decoder. Stage 0 uses a transformer block which keeps the feature resolution unchanged, while stage 1 and 2 employ the proposed UP-Transformer block which enlarges the spatial resolution by $2\times$. The Attention Message Passing (AMP) enables cross-scale self-attention interaction, and the Encoder Feature aggregation incorporates multi-scale information from the InvPT encoder.  ``$\oplus$'' and ``Proj'' denote the accumulation operation and the linear projection layer, respectively.
\textbf{(b)} Pipeline of task-specific reshaping and upsampling block.}
\label{fig:invpt_reshapeup}
\vspace{-18pt}
\end{figure}

\subsection{Task-Specific Preliminary Decoders}\label{sec:preliminarydecoders}
To learn task-specific representations for different tasks, we construct a decoding block consisting of a $3\times 3$ convolutional layer, a batch normalization layer, and a ReLU activation function (\ie~``Conv-BN-ReLU''). The preliminary decoder uses two such blocks to produce task-specific feature for each task.
Then suppose that we have $T$ tasks in total, for the $t$-th task, the output task-specific feature $\mathbf{F}_t^d$ from the preliminary decoder is projected by a linear projection layer (\ie~$1\times 1$ convolution) to produce a preliminary task prediction $\mathbf{P}_t$, which is supervised by the ground-truth labels.
Then, we concatenate $\mathbf{F}_t^{d}$ and $\mathbf{P}_t$ along the channel dimension, and adjust the channel number to $C_0$ with a linear projection layer, to make the different task-specific features with the same channel dimension to facilitate the processing in the transformer decoder.
The combined output is denoted as $\mathbf{F}_t^{c}$.
We flatten $\mathbf{F}_t^c$ spatially to a sequence, and concatenate all $T$ sequences as $\mathbf{F}^{c}$ with $\mathbf{F}^{c} \in \mathbb{R}^{TH_0 W_0 \times C_0}$. $\mathbf{F}^{c}$ serves as the input of the proposed InvPT transformer decoder, which enables global task interaction and progressively generates refined task-specific feature maps with a high-resolution.

\subsection{InvPT Transformer Decoder via UP-Transformer Block}
\label{sec:invpt_decoder}
As global self-attention is prohibitively expensive when the spatial resolution is high, many visual transformer models typically downsample the feature maps dramatically~\cite{vit,cvt,pvt} and output features with low spatial resolution. However, resolution of the feature map is a critical factor for dense prediction problems, as we need to predict task labels for each pixel, and the sizes of semantic objects in images vary tremendously~\cite{pspnet}.
Another point is that different scales of the feature maps can model different levels of visual information~\cite{mti,hrformer}, and thus it is particularly beneficial to make different tasks learn from each other at {multiple scales}. With these motivations, we design a progressively resolution-enlarged transformer decoder termed as ``Inverted Pyramid Transformer Decoder'' (\ie~InvPT decoder), which consists of an efficient UP-Transformer block, cross-scale self-attention message passing, and multi-scale encoder feature aggregation, in a unified network module.

\vspace{3pt}
\par\noindent\textbf{Main Structure} As shown in~Fig.~\ref{fig:invpt_reshapeup}, there are three stages in the InvPT decoder, and each stage is the designed UP-Transformer block computing self-attention and updating features at \emph{different} spatial resolutions. The first stage (\ie~stage 0) of InvPT decoder learns cross-task self-attention at the output resolution of the InvPT encoder (\ie~$H_0 \times W_0$), while
the following two stages enlarge the spatial resolution of the feature maps, and calculate cross-task self-attention at higher resolutions.
The latter two stages (\ie~stage 1 and 2) use the proposed UP-Transformer blocks which refine features at higher resolution and enable cross-scale self-attention propagation as well as multi-scale transformer feature aggregation from the transformer encoder.

\par To simplify the description, for the $s$-th stage (\eg~$s=0,1,2$), we denote its input feature as $\mathbf{F}_s$ with $\mathbf{F}_s \in \mathbb{R}^{TH_s W_s  \times C_s }$, where $H_s$ and $W_s$ are the spatial height and width of the feature, and $C_s$ is the channel number. Thus, the input of stage 0 (\ie~$\mathbf{F}_0$) is $\mathbf{F}^c$, which is a combination of a set of $T$ task-specific features as introduced in Sec.~\ref{sec:preliminarydecoders}.
For the output of InvPT decoder, we add up the feature maps from different stages after the alignment of feature resolution and channel dimension with bilinear interpolation and a linear projection layer, and then pass it to a Conv-BN-ReLU block to produce $T$ upsampled and refined task-specific features for the final prediction.

\vspace{3pt}
\par
\begin{figure*}[t]
\vspace{-10pt}

	\centering
	\includegraphics[width=1.\textwidth]{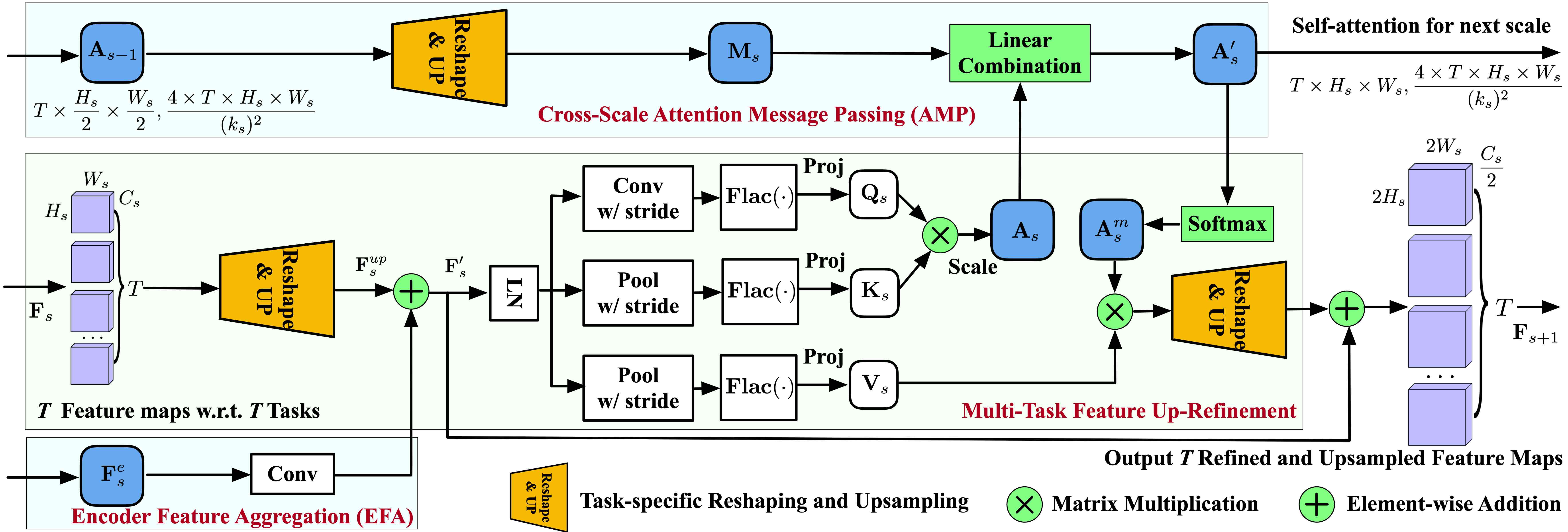}
	\vspace{-15pt}
	\caption{Illustration of the proposed UP-Transformer Block. \textbf{Input}: $\mathbf{A}_{s-1}$ is the attention score matrix of the $(s-1)$-th stage, $\mathbf{F}_s$ is the input multi-task feature sequence of the $s$-th stage, and $\mathbf{F}^e_s$ is the feature passed from the transformer encoder. \textbf{Output}: $\mathbf{A}_s'$ is the enhanced attention score matrix passing to the next stage and $\mathbf{F}_{s+1}$ is the refined and upsampled feature sequence.
}
\vspace{-12pt}
	\label{fig:utb}
\end{figure*}

\noindent\textbf{Task-Specific Reshaping and Upsampling} The transformer computing block typically operates on 2D feature-token sequences with the spatial structure of the feature map broken down, while the spatial structure is critical for dense prediction tasks, and it is not straightforward to directly perform upsampling on the feature sequence with the consideration of spatial structure. Another issue is that the input features into the InvPT decoder consist of multiple different task-specific features, we need to perform feature upsampling and refinement for each single task separately to avoid feature corruption by other tasks. To address these issues, we design a task-specific reshaping and upsampling computation block for the InvPT decoder, \ie~the Reshape \& UP module as illustrated in Fig.~\ref{fig:invpt_reshapeup}~(b) and Fig.~\ref{fig:utb}. {Reshape \& Up} first splits the feature tokens of $\mathbf{F}_s \in \mathbb{R}^{TH_s W_s  \times C_s }$ in the first dimension into $T$ (\ie~the number of tasks) groups of features with tensor slicing, and reshapes each of them back to a spatial feature map with a shape of $\mathbb{R}^{H_s\times W_s\times C_s}$. Then a bilinear interpolation with a scaling factor of 2 is performed to enlarge the height and width by $2\times$. Each scaled task-specific feature map is further fed into a Conv-BN-Relu block to perform feature fusion and reduction of channel dimension if necessary. Finally, The $T$ feature maps are reshaped back to token sequences and finally concatenated as an upsampled multi-task token sequence.
\vspace{3pt}
\par\noindent\textbf{Multi-task UP-Transformer Block} The multi-task UP-Transformer block (see~Fig.~\ref{fig:utb}) is employed in both stage 1 and 2 (\ie~$s=1,2$) and learns to gradually increase the spatial resolution of the multi-task feature and also perform feature interaction and refinement among \emph{all} the tasks in a global manner. As illustrated in Fig.~\ref{fig:utb},
after the Reshape \& Up module with a $3\times 3$ Conv-BN-ReLU computation, we reduce the feature channel dimension by $2\times$ and produce a
learned upsampled feature-token sequence $\mathbf{F}^{up}_s \in \mathbb{R}^{4TH_s W_s \times (C_s/2)}$. $\mathbf{F}_s^{up}$ is first added by the feature sequence passed from the transformer encoder, which we introduce later in this section, and a layer normalization (LN)~\cite{layernorm} is performed on the combined features to produce $\mathbf{F}'_s \in
\mathbb{R}^{4TH_s W_s \times (C_s/2)}$, which serves as the input for the self-attention calculation.

\par To calculate a global self-attention from the upsampled high-resolution multi-task feature sequence $\mathbf{F}'_s$, the memory footprint is prohibitively large. Thus, we first reduce the size of query $\mathbf{Q}_s$, key $\mathbf{K}_s$, and value $\mathbf{V}_s$ matrices for self-attention computation following works~\cite{pvt,cvt}.
Specifically, we first split and reshape the feature-token sequence $\mathbf{F}'_s$ into $T$ groups of spatial feature maps corresponding to the $T$ tasks, each with a shape of $\mathbb{R}^{2H_s\times 2W_s \times (C_s/2)}$. To generate the query embedding from each feature map, we use a convolution with a fixed kernel size $k_c=3$ and a stride $2$ for each group, denoted as $\mathrm{Conv}(\cdot, k_c)$. To generate the key and value embeddings, we use an average pooling operation $\mathrm{Pool}(\cdot, k_s)$ with a kernel size $k_s$ ($k_s = 2^{s+1}$ for the $s$-th stage). By controlling the kernel parameters of the convolution and pooling operations, we can largely improve the memory and computation efficiency of the global self-attention calculation, which makes the utilization of multiple consecutive UP-Transformer blocks possible.
Then, we define $\mathbf{Flac}(\cdot)$ as a function that first flattens the $T$ groups of spatial feature maps and then performs concatenation to produce a multi-task feature-token sequence. Then we can perform global interaction among all the tasks with a multi-task self-attention. Let $\mathbf{W}_s^q$, $\mathbf{W}_s^k$, and $\mathbf{W}_s^v$ be the parameter matrices of three linear projection layers, the calculation of $\mathbf{Q}_s$, $\mathbf{K}_s$, and $\mathbf{V}_s$ can be formulated as follows:
\begin{equation}
\begin{aligned}
\mathbf{Q}_s &= \mathbf{W}^q_s \times \mathbf{Flac}\big(\mathrm{Conv}(\mathbf{F}'_s, k_c)\big), \, \mathbf{Q}_s \in \mathbb{R}^{{TH_s W_s} \times \frac{C_s}{2}}, \\
\mathbf{K}_s &= \mathbf{W}^k_s \times \mathbf{Flac}\big(\mathrm{Pool}(\mathbf{F}'_s, k_s)\big), \, \mathbf{K}_s \in \mathbb{R}^{\frac{4TH_sW_s}{(k_s)^2} \times \frac{C_s}{2}},\\
\mathbf{V}_s &= \mathbf{W}^v_s \times \mathbf{Flac}\big(\mathrm{Pool}(\mathbf{F}'_s, k_s)\big), \, \mathbf{V}_s \in \mathbb{R}^{\frac{4TH_s W_s}{(k_s)^2} \times \frac{C_s}{2}}.
\end{aligned}
\end{equation}

With $\mathbf{Q}_s$ and $\mathbf{K}_s$, the self-attention score matrix $\mathbf{A}_s$ for the $s$-th stage can be calculated as:
\begin{equation} \label{equ:asco}
    \mathbf{A}_s = \frac{\mathbf{Q}_s \mathbf{K}^T_s}{\sqrt{C'_s}}, \, \mathbf{A}_s \in \mathbb{R}^{T H_s W_s \times \frac{4T H_s W_s}{(k_s)^2}},
\end{equation}
where $\sqrt{C'_s}$ with $C'_s=\frac{C_s}{2}$ is a scaling factor to address the magnitude problem~\cite{transformer}. In vanilla transformers, the self-attention map is directly calculated with a softmax function on $\mathbf{A}_s$.~We propose a mechanism of Attention Message Passing (AMP) to enhance the attention $\mathbf{A}_s$ before the Softmax normalization, using multi-scale information from different transformer stages.

\vspace{3pt}
\par\noindent\textbf{Cross-Scale Self-Attention Message Passing} To enable the InvPT decoder to model the cross-task interaction at different scales more effectively, we pass attention message to the attention score matrix $\mathbf{A}_s$ at the current $s$-th stage from $\mathbf{A}_{s-1} \in \mathbb{R}^{\frac{T H_s W_s}{4} \times \frac{4T H_s W_s}{(k_s)^2}}$, which is computed at the $(s-1)$-th stage. It can be noted that the second dimension of $\mathbf{A}_s$ at different stages maintain the same size due to the design of the kernel size $k_s$.
An illustration of AMP is shown in~Fig.~\ref{fig:utb}. Specifically, we perform `Reshape \& Up' operation to first adjust the shape of $\mathbf{A}_{s-1}$ to $\mathbb{R}^{\frac{H_s}{2}\times \frac{W_s}{2} \times \frac{TH_sW_s}{(k_s)^2}}$, and then perform $2\times$ bilinear interpolation in the first two spatial dimensions, and finally flatten it to have the attention message matrix $\mathbf{M}_{s-1} \in \mathbb{R}^{{T H_s W_s} \times \frac{4T H_s W_s}{(k_s)^2}}$ which has the same dimension as $\mathbf{A}_s$.
Finally, we perform a linear combination of the self-attention maps $\mathbf{A}_s$ and $\mathbf{M}_{s-1}$ to pass attention message from the $(s-1)$-th stage to the $s$-th stage as:
\begin{equation}\label{equ_attnMap}
\begin{aligned}
\mathbf{A}'_{s} &=  \alpha^1_s \mathbf{A}_{s} +  \alpha^2_s \mathbf{M}_{s-1},\\
\mathbf{A}_s^{m} &= \mathrm{Softmax} (\mathbf{A}'_s),
\end{aligned}
\end{equation}
where $\alpha^1_s$ and $\alpha^2_s$ are learnable weights for $\mathbf{A}_s$ and $\mathbf{M}_{s-1}$, respectively;
$\mathrm{Softmax(\cdot)}$ is a \emph{row-wise} softmax function. After multiplying with the new attention map $\mathbf{A}^{m}_s$ with the value matrix $\mathbf{V}_s$, we obtain the final multi-task feature $\mathbf{F}_{s+1}$ as output, which is refined and upsampled, and can be further fed into the next UP-Transformer block. The whole process can be formulated as follows:
\begin{equation}
\mathbf{F}_{s+1} = \text{Reshape\_Up}(\mathbf{A}^{m}_s \times \mathbf{V}_s) + \mathbf{F}'_s,
\end{equation}
where the function $\text{Reshape\_Up}(\cdot)$ denotes the task-specific reshaping and upsampling. It enlarges the feature map by $2\times$ to be the same spatial resolution as $\mathbf{F}'_s$ which is a residual feature map from the input.

\par\noindent\textbf{Efficient Multi-Scale Encoder Feature Aggregation}
For dense scene understanding, some basic tasks such as object boundary detection require lower-level visual representation. However, it is tricky to efficiently use the multi-scale features in transformer as standard transformer structures have a quadratic computational complexity regarding the image resolution, and typically only operate on small-resolution feature maps. To gradually increase the feature map size and also incorporate multi-scale features is very challenging to GPU memory.
Therefore, we design an efficient yet effective multi-scale encoder feature aggregation (EFA) strategy. As shown in~Fig.~\ref{fig:utb},
in each upsampling stage~$s$ of the InvPT decoder, we obtain a corresponding-scale feature sequence $\mathbf{F}^e_s$ with channel number $C_s^e$ from the transformer encoder. We reshape it to a spatial feature map and apply a $3\times 3$ convolution to produce a new feature map with channel dimension $C_s$, and then reshape and expand it $T$ times to align with the dimension of $\mathbf{F}_s^{up}$ and add to it. Benefiting from the proposed efficient UP-Transformer block, we can still maintain high efficiency even if at each consecutive stage of the decoder, the resolution is gradually enlarged by two times.

\begin{figure}[!t]
	\centering
	\vspace{-5pt}
	\includegraphics[width=0.8\textwidth]{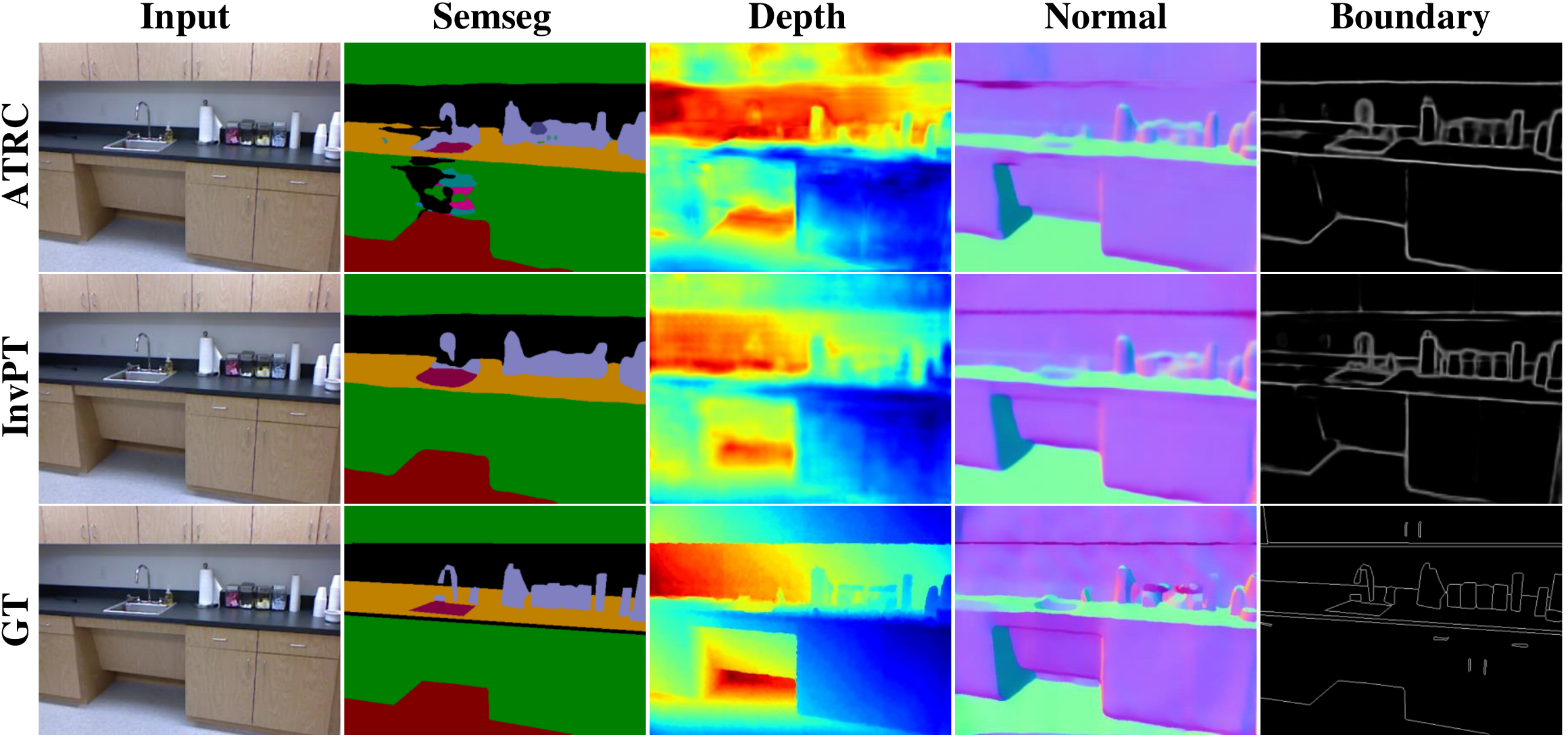}
	\vspace{-0.8em}	
	\caption{Qualitative comparison with the previous best method ATRC~\cite{atrc} on NYUD-v2. Ours produces more accurate predictions on the different tasks.
	}
	\vspace{-20pt}
	\label{fig:qualitative_sota_nyud}
\end{figure}

\vspace{-10pt}
\section{Experiments}
\vspace{-6pt}
We present extensive experiments to demonstrate the effectiveness of the proposed InvPT framework for multi-task dense prediction.
\vspace{3pt}
\par\noindent\textbf{Datasets} The experiments are conducted on two popular scene understanding datasets with multi-task labels, \ie~NYUD-v2~\cite{silberman2012indoor} and PASCAL-Context~\cite{chen2014detect}.
\textbf{NYUD-v2} contains various indoor scenes such as offices and living rooms with 795 training and 654 testing images. It provides different dense labels, including semantic segmentation, monocular depth estimation, surface normal estimation  and object boundary detection.
\textbf{PASCAL-Context} is formed from PASCAL dataset~\cite{everingham2010pascal}. It has 4,998 images in the training split and 5,105 in the testing split, covering both indoor and outdoor scenes. This dataset provides pixel-wise labels for semantic segmentation, human parsing and object boundary detection. Additionally,~\cite{astmt} generates surface normal and saliency labels for this dataset.
We perform experiments on \emph{all} tasks in both datasets for evaluation.

\vspace{3pt}
\par\noindent\textbf{Evaluation}
Semantic segmentation (Semseg) and human parsing (Parsing) are evaluated with mean Intersection over Union (mIoU); monocular depth estimation (Depth) is evaluated with Root Mean Square Error (RMSE);
surface normal estimation (Normal) is evaluated by the mean error (mErr) of predicted angles; saliency detection (Saliency) is evaluated with maximal F-measure (maxF); object boundary detection (Boundary) is evaluated with the optimal-dataset-scale F-measure (odsF). To evaluate the average performance gain of multi-task models against single-task models, we adopt the ``multi-task learning performance'' (MT Performance) metric $\Delta_m$ introduced in~\cite{astmt}, which is an important metric to reflect the performance of a multi-task model (the higher the better).

\vspace{3pt}
\par\noindent\textbf{Implementation Details}
For the ablation study, we adopt Swin-Tiny transformer~\cite{swin} pre-trained on ImageNet-22K~\cite{deng2009imagenet} as the transformer encoder. The models for different evaluation experiments are trained for 40,000 iterations on both datasets, with a batch size of 6. For the transformer models, Adam optimizer is adopted with a learning rate of $2\times 10^{-5}$, and a weight decay rate of $1\times 10^{-6}$. Polynomial learning rate scheduler is used.
The output channel number of preliminary decoder is 768. More details are in Appendix.

\begin{table}[!t]
\centering
\caption{Ablation study on the InvPT decoder. The proposed InvPT and its components yield consistent improvement on different datasets and achieve clear overall improvement on each single task and the multi-task (MT) performance $\Delta_m$. The gain shows absolute performance point improvement. The different variants of the InvPT all use Swin-tiny as its encoder structure.  `$\mathbf{\downarrow}$' means lower better and `$\mathbf{\uparrow}$' means higher better.}
\label{tab:abl_modules_nyudpascal}
\resizebox{1.\linewidth}{!}{
    \begin{tabular}{l|cccc|cccccccc}
    \toprule
        \multicolumn{1}{c|}{ \multirow{3}*{ \textbf{Model}} }   & \multicolumn{4}{c|}{\textbf{NYUD-v2}} & \multicolumn{5}{c}{\textbf{PASCAL-Context}} \\
        \cline{2-10}
        & \textbf{Semseg}  & \textbf{Depth}  & \textbf{Normal} & \textbf{Boundary} & \textbf{Semseg}  & \textbf{Parsing}  & \textbf{Saliency} & \textbf{Normal} & \textbf{Boundary}  \\
        &mIoU $\mathbf{\uparrow}$ & RMSE $\mathbf{\downarrow}$ & mErr $\mathbf{\downarrow}$ & odsF $\mathbf{\uparrow}$ & mIoU $\mathbf{\uparrow}$  & mIoU $\mathbf{\uparrow}$
       & maxF $\mathbf{\uparrow}$ & mErr $\mathbf{\downarrow}$ & odsF $\mathbf{\uparrow}$ \\
    \midrule
    InvPT Baseline (ST)& 43.29& 0.5975 & 20.80 & 76.10 & 72.43 & 61.13 & 83.43 & 14.38 &  71.50 \\
     \midrule
    InvPT Baseline (MT) & 41.06  & 0.6350 & 21.47& 76.00 & 70.92 & 59.63 & 82.63 & 14.63 & 71.30\\
    InvPT w/ UTB  & 43.18 & 0.5643 & 21.05 & 76.10 & 72.34 & 61.08 & 83.99 & 14.49 & 71.60  \\
    InvPT w/ UTB+AMP &  43.64 & 0.5617 & 20.87 & 76.10 & 73.29 & 61.78 & 84.03 & 14.37 & 71.80 \\
     InvPT w/ UTB+AMP+EFA (Full) & 44.27 & 0.5589 & 20.46 & 76.10 & 73.93 & 62.73 & 84.24 & 14.15& 72.60\\
     Gain on Each Task (vs. MT) & $\bigtriangleup$\textbf{3.21} &$\bigtriangleup$ \textbf{0.0761} &$\bigtriangleup$ \textbf{1.01} & $\bigtriangleup$\textbf{0.10} & $\bigtriangleup$\textbf{3.01} & $\bigtriangleup$\textbf{3.10} &$\bigtriangleup$ \textbf{1.61} & $\bigtriangleup$\textbf{0.48} &$\bigtriangleup$ \textbf{1.30}\\
     \midrule
     {MT Performance $\Delta_m$ (vs. ST)}~\cite{astmt}& \multicolumn{4}{c|}{+\textbf{2.59}} & \multicolumn{5}{c}{+\textbf{1.76}} \\
    \bottomrule
    \end{tabular}}
     \vspace{-10pt}
\end{table}

\begin{figure}[!t]
	\centering
	\includegraphics[width=0.98\textwidth]{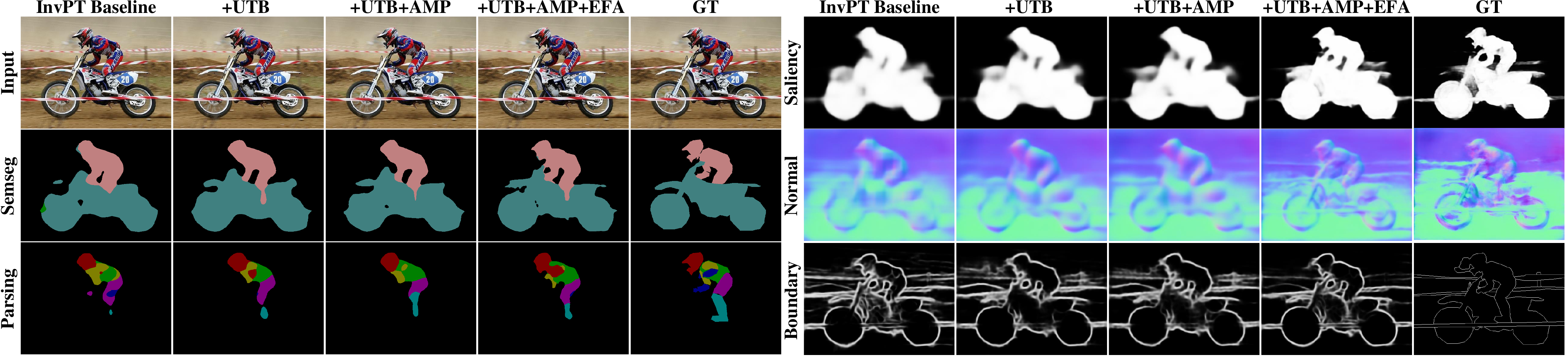}
    \vspace{-8pt}
	\caption{Qualitative analysis of InvPT decoder on PASCAL-Context. Results of different model variants are shown by columns.
	}
	\label{fig:qualitative_ablation_2}
\vspace{-18pt}
\end{figure}

\subsection{Model Analysis}
\par\noindent\textbf{Baselines and Model Variants} To have a deep analysis of the proposed InvPT framework, we first define several model baselines and variants (see Table~\ref{tab:abl_modules_nyudpascal}):
(i) {``InvPT Baseline (MT)''} denotes a strong multi-task baseline model of the proposed  InvPT framework. It uses Swin-tiny encoder and two $3\times 3$ Conv-BN-ReLU blocks as decoder for each task, which is equivalent to the preliminary decoder in InvPT. The encoder feature map is upsampled by $8\times$ before the final prediction. It also combines multi-scale features from the encoder to help boost performance. This is a typical multi-task baseline structure as in previous works~\cite{mti,atrc}.
(ii) ``InvPT Baseline (ST)'' has a similar structure as ``InvPT Baseline'' but it is trained under single-task setting. (iii) ``InvPT w/ UTB'' indicates adding the proposed UP-Transformer block upon ``InvPT Baseline (MT)''; Similarly, ``InvPT w/ UTB + AMP'' indicates further adding the cross-scale Attention Message Passing, and ``InvPT w/ UTB + AMP + EFA'' denotes the full model by further adding multi-scale Encoder Feature Aggregation.

\begin{figure*}[!t]
\begin{minipage}[c]{1\textwidth}
\begin{minipage}[c]{0.45\textwidth}
\begin{subfigure}[t]{1\linewidth}
\centering
\includegraphics[width=\textwidth]{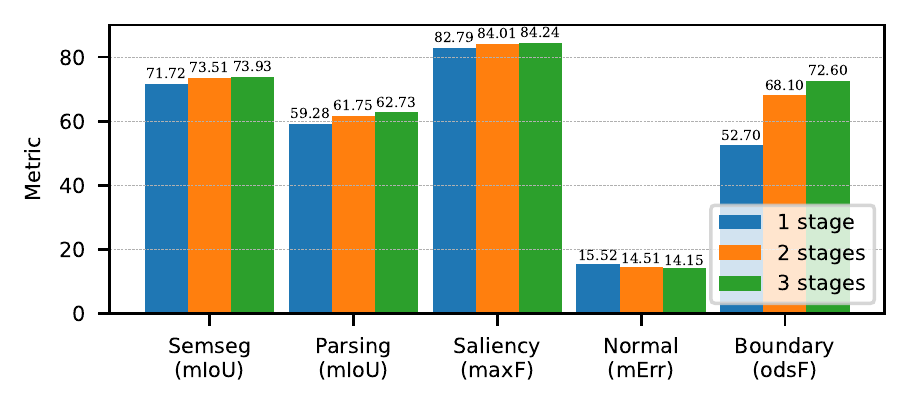}
\end{subfigure}
\vspace{-18pt}
\figcaption{Investigation of the number of stages in the InvPT decoder.}
\label{fig:stages}
\end{minipage}
\hspace{3pt}
\begin{minipage}[c]{0.53\textwidth}
\centering
\resizebox{1.\linewidth}{!}{
 \setlength{\tabcolsep}{3mm}{
    \begin{tabular}[t]{lccccc}
    \toprule
   \multirow{2}*{ \textbf{Encoder} }  & \textbf{Semseg}  & \textbf{Parsing}
       & \textbf{Saliency} & \textbf{Normal} & \textbf{Boundary}
     \\
        & mIoU $\mathbf{\uparrow}$  & mIoU $\mathbf{\uparrow}$
       & maxF $\mathbf{\uparrow}$ & mErr $\mathbf{\downarrow}$ & odsF $\mathbf{\uparrow}$
     \\
    \midrule
     Swin-T & 73.93 & 62.73 & \textbf{84.24} & \textbf{14.15} & 72.60\\
     Swin-B  & 77.50 & 66.83 & 83.65 & 14.63 & 73.00   \\
     Swin-L & \textbf{78.53} & \textbf{68.58} & 83.71 & 14.56 & \textbf{73.60}   \\
     \midrule
     Vit-B & 77.33 & 66.62 & \textbf{85.14} & \textbf{13.78} & \textbf{73.20} \\
     Vit-L & \textbf{79.03} & \textbf{67.61} & 84.81 & 14.15 & 73.00\\
    \bottomrule
   \end{tabular}}}
    \vspace{-3.5pt}

\captionof{table}{Performance comparison of using different transformer encoder structures in InvPT on PASCAL-Context.}
\label{tab:abl_encoder_pascal}
\end{minipage}
\end{minipage}
\vspace{-10pt}
\end{figure*}

\par\noindent\textbf{Effectiveness of InvPT Decoder}
In this part, we investigate the effectiveness of the proposed three modules to demonstrate the proposed InvPT decoder, \ie~UP-Transformer Block (UTB), cross-scale Attention Message Passing (AMP), and Encoder Feature Aggregation (EFA), on both datasets. The experimental results for this investigation are shown in~Table~\ref{tab:abl_modules_nyudpascal}. It can be observed that the UTB, AMP and EFA modules all achieve clear improvement. Specifically, as a core module of the proposed InvPT framework, UTB significantly improves the task Semseg by 2.12 (mIoU), Depth by 0.0707 (RMSE) and Normal by 0.42 (mErr) on NYUD-v2.
Finally, the full model of InvPT achieves remarkable performance gain compared against both the Single-task (ST) baseline (see MT performance $\Delta_m$) and the multi-task (MT) baseline (see the Gain on each task), clearly verifying the effectivenss of the proposed InvPT decoder.
\par For qualitative comparison, in Fig.~\ref{fig:qualitative_ablation_2}, we show prediction examples generated by different model variants which add these modules one by one on PASCAL-Context. It is intuitively to observe that the proposed UTB, AMP and EFA all help produce visually more accurate predictions to against the baseline.

\vspace{3pt}
\par\noindent\textbf{Multi-task Improvement against Single-task Setting}
To validate the effectiveness of the proposed multi-task model, we compare it with its single-task variant ``InvPT Baseline (ST)'' on both datasets in~Table~\ref{tab:abl_modules_nyudpascal}.
Our full model achieves strong performance improvement against the single-task model, yielding $\textbf{2.59\%}$ multi-task performance on NYUD-v2 and $\textbf{1.76\%}$ on PASCAL-Context.

\begin{figure}[t]
\centering
\vspace{-5pt}
\begin{subfigure}[t]{\textwidth}
    \centering
    \includegraphics[width=0.98\textwidth]{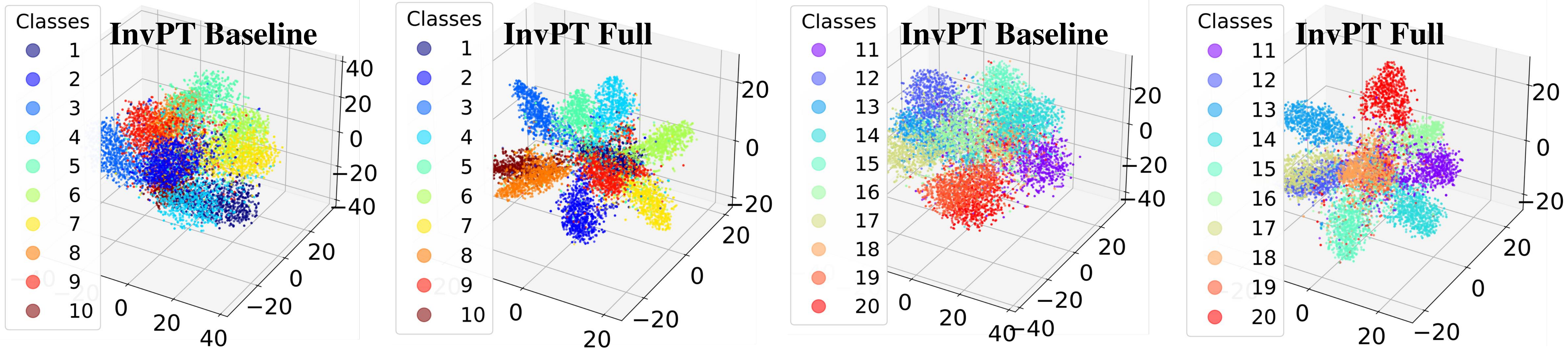}
    \caption{Statistics of learned features with t-SNE~\cite{van2008visualizing} of all 20 classes on Pascal-Context.}
    \label{fig:tsne}
\end{subfigure}

\begin{subfigure}[t]{\textwidth}
    \centering
    \includegraphics[width=0.98\textwidth]{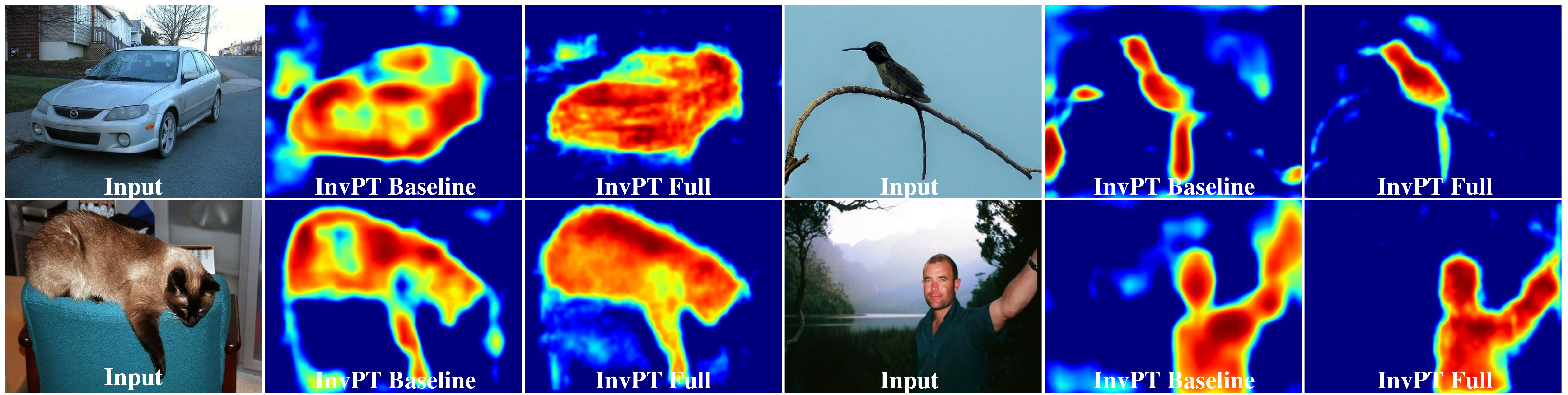}
    \caption{Visualization of examples of learned feature from the semantic segmentation task.}
    \label{fig:featimprove}
\vspace{-8pt}
\end{subfigure}
\caption{Qualitative interpretation of learned features using the proposed InvPT method. From both (a) and (b), it can be observed that features learned by InvPT is effectively improved and is more discriminative compared to the baseline.}
\label{fig:tsne_featimprove}
\vspace{-18pt}
\end{figure}

\par\noindent\textbf{Effect of Different Transformer Encoders}
We also compare the model performance using two families of transformer encoders, \ie~Swin Transformer (Swin-T, Swin-B, and Swin-L)~\cite{swin} and ViT (ViT-B and ViT-L)~\cite{vit}, used as our InvPT encoder. The results on PASCAL-Context are shown in~Table~\ref{tab:abl_encoder_pascal}, and results on NYUD-v2 can be found in Appendix. We observe that the models with higher capacity in the same family can generally obtain consistent performance gain on tasks including Semseg and Parsing, while on other lower-level tasks (\eg~Boundary) the improvement is however not clear.
One possible reason for the difference of performance gain is the task competition problem in training as discussed in previous works~\cite{sener2018multi,kendall2018multi}.

\par\noindent\textbf{Effect of the Number of Stages} Our UP-Transformer block typically consists of three stages. In~Fig.~\ref{fig:stages}, we show how the number of stages of InvPT decoder influences the performance of the different tasks on PASCAL-Context. It can be observed that using more stages can help the InvPT decoder learn better predictions for all tasks. Our efficient design makes the multi-task decoding feature maps with gradually increased resolutions possible.

\par\noindent{\textbf{Qualitative Study of learned Features with InvPT}}
In Fig.~\ref{fig:tsne_featimprove}, we show visualization comparison of the learned final features between the transformer baseline (\ie~InvPT Baseline (MT)) and our InvPT full model, to further demonstrate how the features are improved using our proposed InvPT model.
The statistics of the learned feature points is visualized with t-SNE~\cite{tsne} on all 20 semantic classes of Pascal-Context dataset. It is obvious that our model helps learn more discriminative features, thus resulting in higher quantitative results. The generated spatial feature maps for segmentation are also intuitively better.

\par
\begin{figure}[!t]
	\centering
	\includegraphics[width=0.98\textwidth]{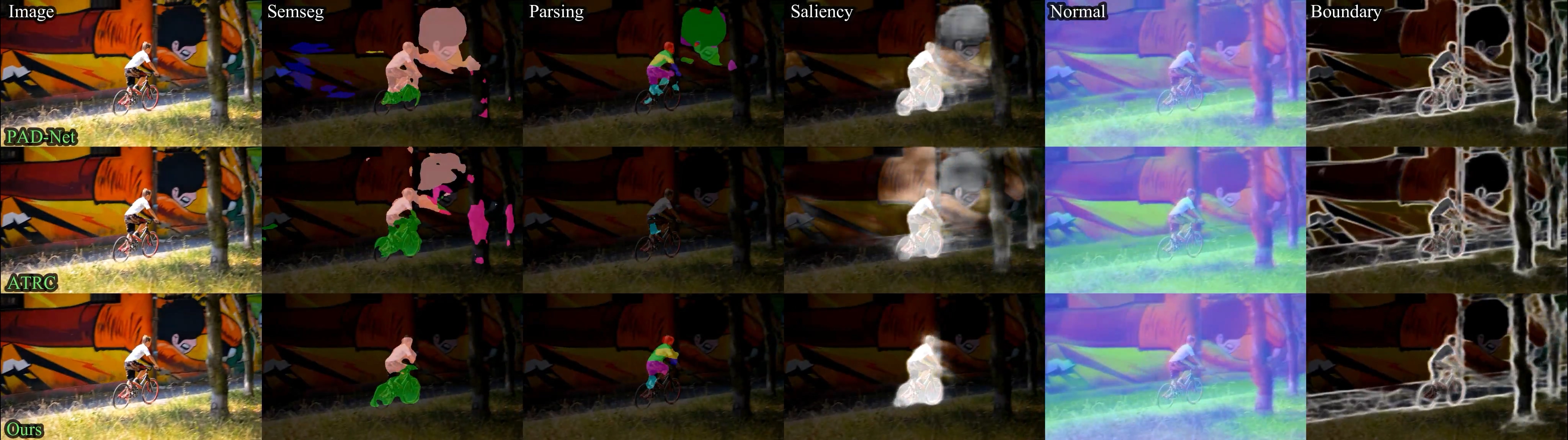}
	\vspace{-8pt}
    \caption{Study on generalization performance. Models are all trained on PASCAL-Context~\cite{chen2014detect} and tested on DAVIS video dataset~\cite{davis}. The proposed method yields better generalization performance compared to PAD-Net~\cite{padnet} and ATRC~\cite{atrc}.
 	}
 	\label{fig:demo}
	\vspace{-15pt}
\end{figure}

\noindent\textbf{Generalization Performance}
To qualitatively study the generalization performance of the proposed multi-task transformer for dense scene understanding, we compare it with the best performing methods, including ATRC~\cite{atrc} and PAD-Net~\cite{padnet}, on the challenging DAVIS video segmentation Dataset~\cite{davis}. The results are shown in the video demo in the github page. All the models are trained on PASCAL-Context with all the five tasks.
Then the models are directly tested on DAVIS to generate multi-task predictions on video sequences. One example frame is shown in Fig.~\ref{fig:demo}, which clearly shows our advantage on this perspective.

\subsection{State-of-the-art Comparison}
Table~\ref{tab:sota_s} shows a comparison of the proposed InvPT method against existing state-of-the-arts, including PAD-Net~\cite{padnet}, MTI-Net~\cite{mti} and ATRC~\cite{atrc}, on both NYUD-v2 and PASCAL-Context.
On all the 9 metrics from these two benchmarks, the proposed InvPT achieves clearly superior performance, especially for higher-level scene understanding tasks such as Semseg and Parsing.
Notably, on NYUD-v2, our InvPT surpasses the previous best performing method (\ie~ATRC) by~\textbf{+7.23} (mIoU) on Semseg,
while on PASCAL-Context, InvPT outperforms ATRC by~\textbf{+11.36}~(mIoU) and \textbf{+4.68}~(mIoU) on Semseg and Parsing, respectively.
A qualitative comparison with ATRC is shown in Fig.~\ref{fig:qualitative_sota_nyud}.

\begin{table}[t]
\centering\caption{State-of-the-art comparison on NYUD-v2 (\emph{left}) and PASCAL-Context (\emph{right}). Our InvPT significantly outperforms the previous state-of-the-arts by a large margin. `$\mathbf{\downarrow}$' means lower better and `$\mathbf{\uparrow}$' means higher better.}
\vspace{3pt}
\label{tab:sota_s}
\begin{minipage}[c]{\textwidth}
\begin{subtable}[t]{0.461\textwidth}
\centering
\resizebox{1.\linewidth}{!}{
\setlength{\tabcolsep}{1.5mm}{
    \begin{tabular}{lcccccc}
    \toprule
         \multirow{2}*{ \textbf{Model} }   & \textbf{Semseg}  & \textbf{Depth}  & \textbf{Normal} & \textbf{Boundary} \\
         &mIoU $\mathbf{\uparrow}$ & RMSE $\mathbf{\downarrow}$ & mErr $\mathbf{\downarrow}$ & odsF $\mathbf{\uparrow}$ \\
    \midrule
    Cross-Stitch~\cite{crossstitch} & 36.34 & 0.6290 & 20.88 &76.38    \\
    PAP~\cite{papnet} & 36.72 & 0.6178 &20.82 & 76.42 \\
    PSD~\cite{psd} & 36.69 & 0.6246 & 20.87 & 76.42 \\
    PAD-Net~\cite{padnet} & 36.61 & 0.6270 & 20.85 & 76.38\\
    MTI-Net~\cite{mti} & 45.97  & 0.5365 & 20.27 & 77.86 \\
    ATRC~\cite{atrc}  & 46.33  & 0.5363 & 20.18 & 77.94 \\
    InvPT (\textbf{ours}) &\textbf{53.56} &\textbf{0.5183} & \textbf{19.04} & \textbf{78.10}\\
    \bottomrule
    \end{tabular}}}
    \label{tab:sota_nyud}
\end{subtable}
\begin{subtable}[t]{0.531\textwidth}
\centering
\resizebox{1.0\linewidth}{!}{
\setlength{\tabcolsep}{1mm}{
    \begin{tabular}{lcccccc}
    \toprule
   \multirow{2}*{ \textbf{Model} }  & \textbf{Semseg}  & \textbf{Parsing}
       & \textbf{Saliency} & \textbf{Normal} & \textbf{Boundary}
     \\
        & mIoU $\mathbf{\uparrow}$  & mIoU $\mathbf{\uparrow}$
       & maxF $\mathbf{\uparrow}$ & mErr $\mathbf{\downarrow}$ & odsF $\mathbf{\uparrow}$
     \\
    \midrule
    ASTMT~\cite{astmt} & 68.00 & 61.10 & 65.70 &  14.70 & 72.40\\
    PAD-Net~\cite{padnet} & 53.60 & 59.60 & 65.80 & 15.30 & 72.50 \\
    MTI-Net~\cite{mti}  &  61.70 & 60.18 & 84.78 & 14.23 & 70.80\\
    ATRC~\cite{atrc}  & 62.69 & 59.42 & 84.70 & 14.20 & 70.96\\
    ATRC-ASPP~\cite{atrc}  & 63.60 & 60.23 & 83.91 & 14.30 & 70.86\\
    ATRC-BMTAS~\cite{atrc}  & 67.67 & 62.93 & 82.29 & 14.24 & 72.42\\
    InvPT (\textbf{ours}) & \textbf{79.03} & \textbf{67.61} & \textbf{84.81} & \textbf{14.15} & \textbf{73.00}\\
    \bottomrule
    \end{tabular}}}
    \label{tab:sota_pascal}
\end{subtable}
\end{minipage}
\vspace{-20pt}
\end{table}

\section{Conclusion}
This paper presented a novel transformer framework, Inverted Pyramid Multi-task Transformer (InvPT), for the multi-task  dense prediction for visual scene understanding.
InvPT is able to effectively learn the long-range interaction in both spatial and all-task contexts on the multi-task feature maps with gradually increased spatial resolution for dense prediction.
Extensive experiments demonstrated the effectiveness of the proposed method, and also showed its significantly better performance on two popular benchmarks compared to the previous state-of-the-art methods.

\par \noindent \textbf{Acknowledgements.}
This research is supported in part by the Early Career Scheme of the Research Grants Council (RGC) of the Hong Kong SAR under grant No. 26202321 and HKUST Startup Fund No. R9253.

\clearpage
\bibliographystyle{splncs04}
\bibliography{refers}
\clearpage

\appendix
\section{Appendix}
\subsection{More Implementation Details}
In this section, we provide more details about our model implementation in addition to those discussed in the paper. 

\par\noindent\textbf{Model Optimization.} For evaluation on NYUD-v2~\cite{silberman2012indoor} and PASCAL-Context~\cite{chen2014detect}, we totally consider six dense prediction tasks, including
semantic segmentation (Semseg), monocular depth estimation (Depth),  surface normal estimation (Normal), human parsing (Parsing), saliency detection (Saliency), and object boundary detection (Boundary).
For the continuous regression tasks (\ie~Depth and Normal)
a $\mathcal{L}1$ Loss is employed. For the discrete classification tasks (\ie~Semseg, Parsing, Saliency, and Boundary), a cross-entropy loss is utilized.
For the sake of simplicity, we use the same set of loss functions for both intermediate and final supervision.
The whole model can be end-to-end optimized.

\par\noindent\textbf{Data Processing.} For a fair comparison with ATRC~\cite{atrc}, we follow its data processing pipeline. On PASCAL-Context, we pad the image to the size of $512\times 512$, while on NYUD-v2, we randomly crop the input image to the size of $448\times 576$ as Swin Transformer~\cite{swin} requires both the height and width to be even for conducting patch merging. We use typical data augmentation including random scaling, cropping, horizontal flipping and color jittering.

\par\noindent\textbf{Implementation Details of Encoder Feature Aggregation (EFA).}
For Swin Transformer encoders~\cite{swin}, we pass feature sequences from the first three stages to Inverted Pyramid Transformer Decoder (InvPT decoder).
For ViT encoders~\cite{vit}, since they do not explicitly define the concept of stage, we evenly choose 3 layers based on the depth and unfold their output spatially, and then use transposed convolution to upsample the resolution of feature maps to match the spatial resolution in the corresponding decoder stage before further transformation.
Specifically, for ViT-base encoder, we use the output token sequences of layer 3, 6, and 9, while for ViT-large encoder we use output token sequences of layer 6, 12, and 18.
The kernel size and stride of the transposed convolution for the feature at the first scale are 4, and those at the second scale are 2.

\par\noindent\textbf{Details about Self-attention in InvPT Decoder.} The specific shapes of the query, the key, and the value matrices (\ie~$\mathbf{Q}, \mathbf{K}, \mathbf{V}$) in different UP-Transformer stages are shown in Table~\ref{tab:qkv}. Please refer to Sec. 3.4 in paper for the detailed definitions of the notations in the table.

\renewcommand{\arraystretch}{1.3}
\begin{table}[ht]
\setlength{\tabcolsep}{4pt}
\vspace{-20pt}
\centering
\caption{Shapes of $\mathbf{Q}, \mathbf{K}, \mathbf{V}$ matrices in different upsampling stages. Please refer to Sec. 3.4 in paper for the detailed definitions of the notations in the table.}
\label{tab:qkv}
\resizebox{0.5\linewidth}{!}{
    \begin{tabular}{cccc}
    \hline
  & $s=0$   & $s=1$  & $s=2$   \\
    \hline
     $\mathbf{Q}$  & $\frac{TH_0 W_0}{4}, C_0$  & $TH_0 W_0, \frac{C_0}{2}$ &  $4TH_0 W_0, \frac{C_0}{4}$\\
     \hline
     $\mathbf{K}$ & $\frac{TH_0 W_0}{4}, C_0$ &  $\frac{TH_0 W_0}{4}, \frac{C_0}{2}$ & $\frac{TH_0 W_0}{4}, \frac{C_0}{4}$\\
      \hline
     $\mathbf{V}$ & $\frac{TH_0 W_0}{4}, C_0$  &  $\frac{TH_0 W_0}{4}, \frac{C_0}{2}$  & $\frac{TH_0 W_0}{4}, \frac{C_0}{4}$\\
     \hline
    \end{tabular}}
\vspace{-2pt}
\end{table}
\renewcommand{\arraystretch}{1}

\subsection{More Experimental Results and Analysis}

\begin{figure*}[t]
	\centering
	\includegraphics[width=1\textwidth]{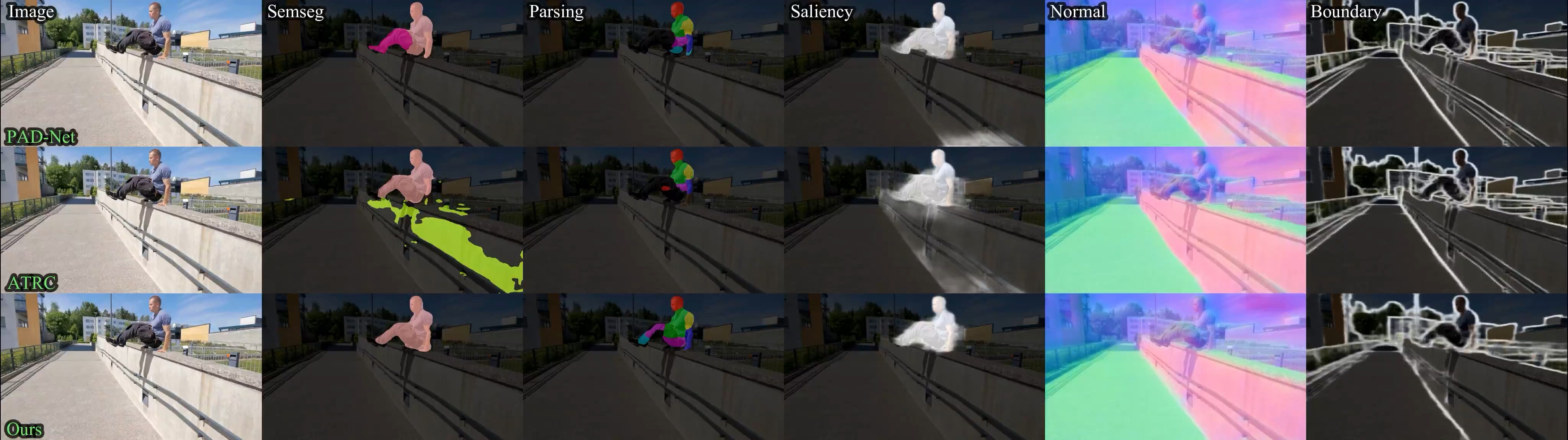}
	\vspace{-15pt}
	\caption{An example frame of the demo video for the study of generalization performance. Models are all trained on PASCAL-Context and tested on DAVIS video dataset. Our method yields qualitatively better generalization performance compared to PAD-Net~\cite{padnet} and ATRC~\cite{atrc}.
	}
	\label{fig:demo2}
	\vspace{-15pt}
\end{figure*}

\par\noindent\textbf{Video Demo for Generalization Performance Comparison on DAVIS Video Dataset.}
As introduced in the paper, to qualitatively study the generalization ability of the proposed multi-task transformer for dense scene understanding, we compare it with the best performing method, including ATRC~\cite{atrc} and PAD-Net~\cite{padnet}, on the challenging DAVIS video Dataset~\cite{davis}.
The results are shown in the attached video demo.
All the models are trained on Pascal-Context with 5 tasks, \ie~semantic segmentation, surface normal estimation, human parsing, saliency detection, and object boundary detection.
Then the models are directly tested on DAVIS to generate multi-task predictions in the demo video.
Significantly stronger generalization ability of our InvPT is observed and
an example frame is shown in Fig.~\ref{fig:demo2}.

\par\noindent\textbf{Effect of Different Transformer Encoders.} Similar to the results on PASCAL-Context in the paper, we compare two families of transformer encoders: Swin Transformer (Swin-T, Swin-B and Swin-L)~\cite{swin} and ViT (ViT-B and ViT-L)~\cite{vit} on NYUD-v2~(Table~\ref{tab:abl_encoder_nyud}).
We observe that the bigger model of the same model family consistently brings performance gain on semantic segmentation and monocular depth estimation, while on other dense tasks (\ie~saliency and boundary detection), it does not necessarily yield significantly better performance despite with higher model capacity. This phenomenon may result from the distinct characteristics of different dense prediction tasks.

\begin{table}[t]
\centering
\caption{Performance comparison of using different transformer
encoder structures in InvPT on NYUD-v2.}
\label{tab:abl_encoder_nyud}
\resizebox{0.5\linewidth}{!}{
    \begin{tabular}{ccccccc}
    \toprule
    \multirow{2}*{ \textbf{Model} }   & \textbf{Semseg}  & \textbf{Depth}  & \textbf{Normal} & \textbf{Boundary} \\
         &mIoU $\mathbf{\uparrow}$ & RMSE $\mathbf{\downarrow}$ & mErr $\mathbf{\downarrow}$ & odsF $\mathbf{\uparrow}$ \\
    \midrule
     Swin-T &   44.27 & 0.5589 & 20.46 & 76.10 \\
     Swin-B  & 50.97 & 0.5071 & \textbf{19.39} & 77.30 \\
     Swin-L &  \textbf{51.76} & \textbf{0.5020} & \textbf{19.39} & \textbf{77.60}\\
     \hline
     Vit-B & 50.30 & 0.5367 & \textbf{19.00} & 77.60 \\
     Vit-L & \textbf{53.56} &\textbf{0.5183} & 19.04 & \textbf{78.10}\\
    \bottomrule
    \end{tabular}}
     \vspace{-5pt}
\end{table}

\par
\begin{figure*}[ht]
	\centering
	\includegraphics[width=0.98\textwidth]{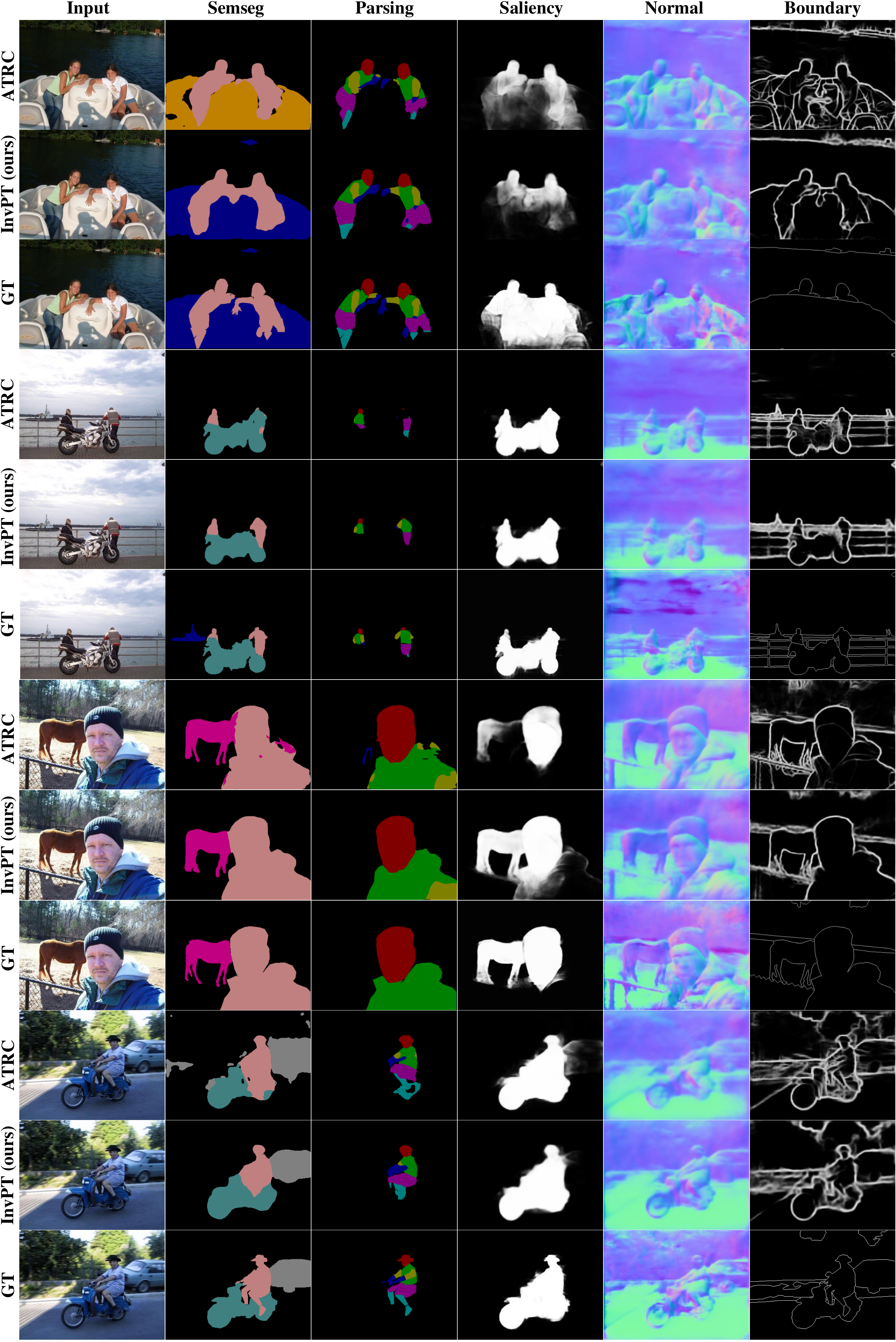}
	\caption{Qualitative comparison with the best performing method ATRC~\cite{atrc} on PASCAL-Context. Our method generates significantly better results especially on semantic segmentation and human parsing.
	}
	\label{fig:qualitative_pascal}
\end{figure*}

\begin{figure*}[ht]
	\centering
	\includegraphics[width=0.98\textwidth]{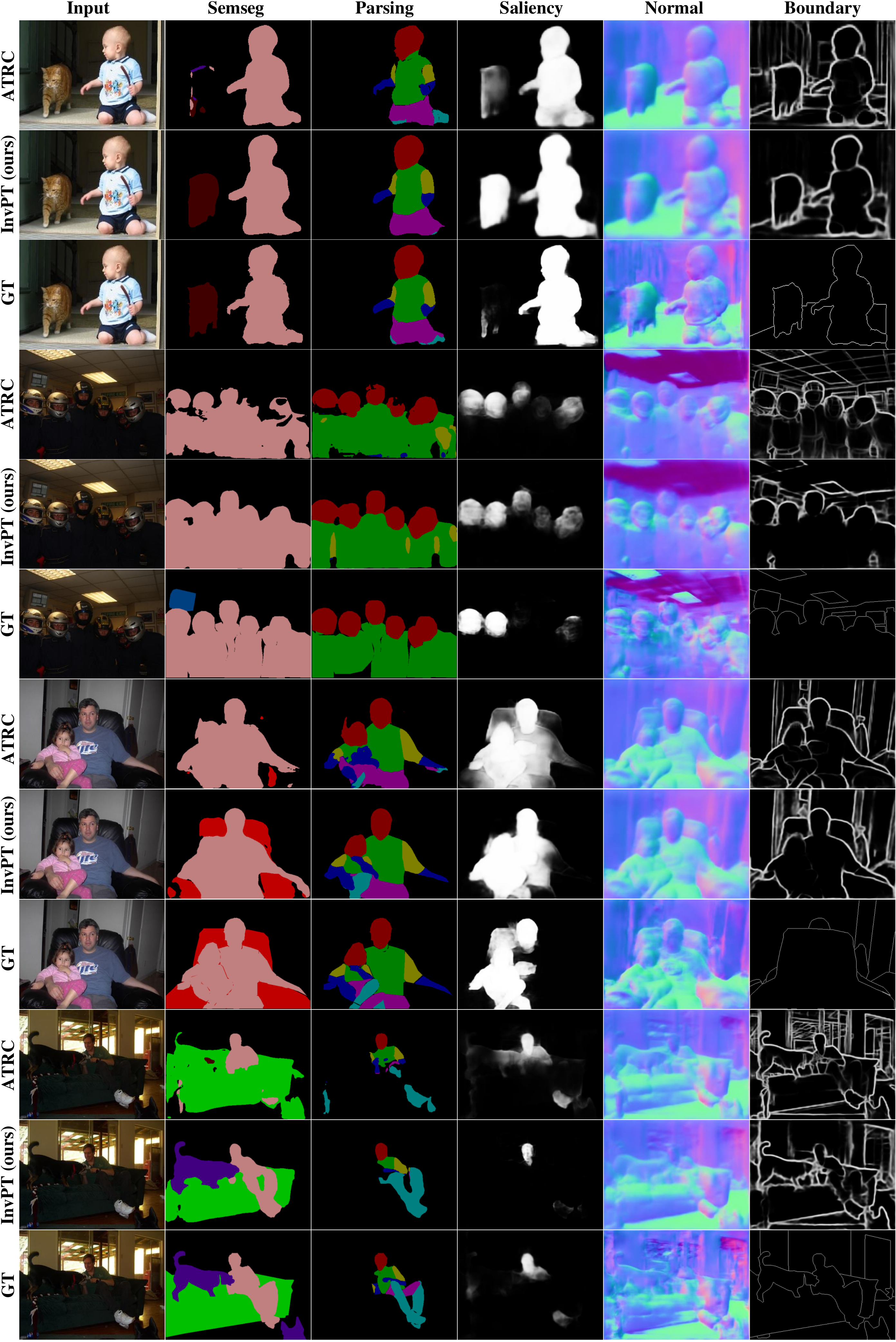}
	\caption{Qualitative comparison with the best performing method ATRC~\cite{atrc} on PASCAL-Context. Our method generates significantly better predictions especially on semantic segmentation and human parsing.
	}
	\label{fig:qualitative_pascal2}
\end{figure*}

\noindent\textbf{More Qualitative Results.}
We show more prediction results by InvPT (ours) and the SOTA method ATRC~\cite{atrc} on the challenging PASCAL-Context dataset in~Fig.~\ref{fig:qualitative_pascal} and Fig.~\ref{fig:qualitative_pascal2}. It is clear that our method produces significantly better results than ATRC, especially on semantic segmentation and human parsing.

\par
\begin{figure*}[ht]
	\centering
	\includegraphics[width=1\textwidth]{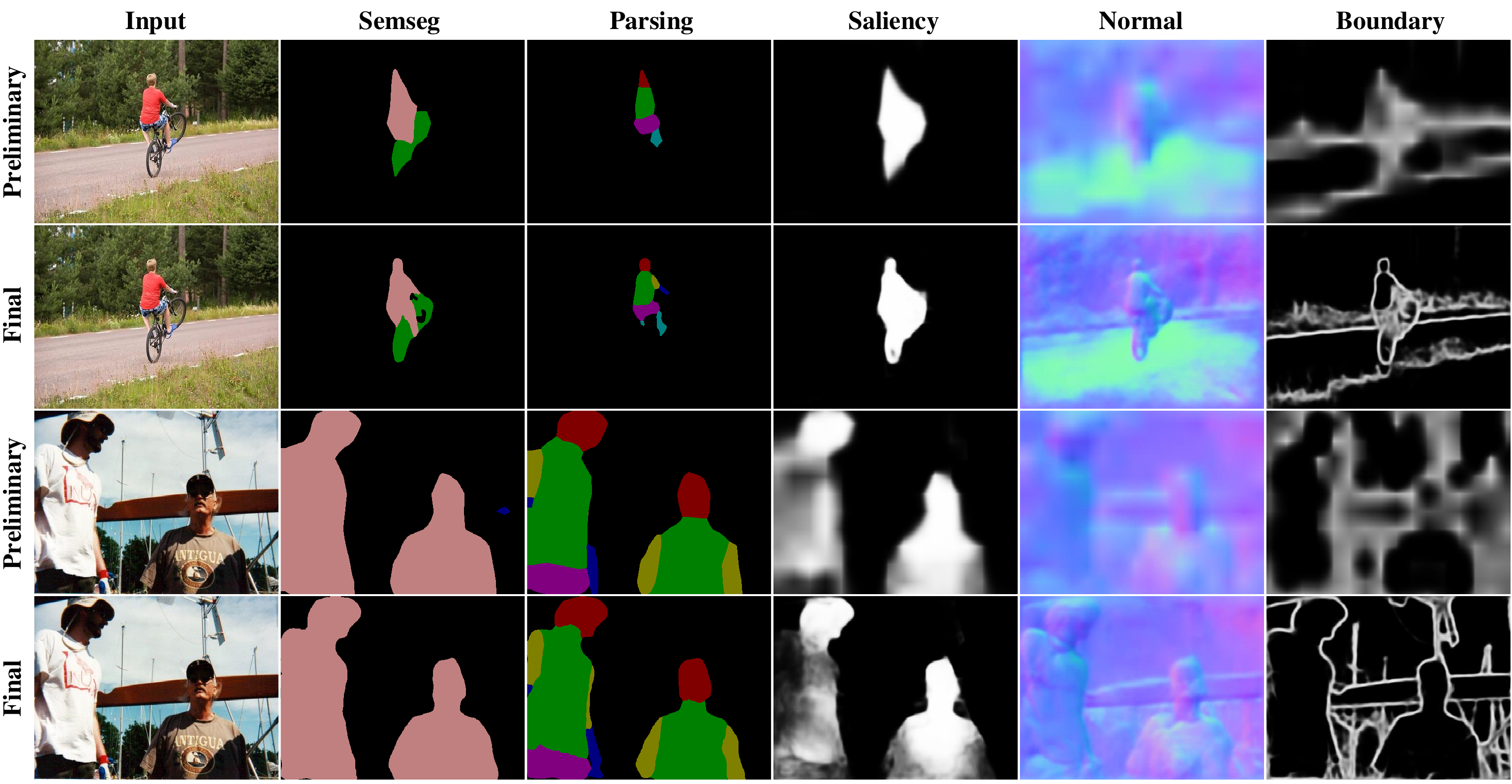}
	\vspace{-15pt}
	\caption{Qualitative comparison of the  predictions from the preliminary decoder and the final predictions of InvPT decoder on PASCAL-Context. The final predictions on all these tasks are significantly more accurate.
	}
	\vspace{-15pt}
	\label{fig:vis_inter}
\end{figure*}

\noindent\textbf{Qualitative Comparison of the Preliminary and Final Predictions of InvPT.}
Fig.~\ref{fig:vis_inter} shows the qualitative comparison of the preliminary predictions and the final predictions generated by InvPT on PASCAL-Context. We can observe that InvPT decoder successfully refines the preliminary predictions and generates remarkably better results on all these dense prediction tasks.

\end{document}

%% file: math_commands.tex
\usepackage{amsmath,amsfonts,bm}

\def\eqref#1{equation~\ref{#1}}

\def\1{\bm{1}}

\newcommand{\test}{\mathcal{D_{\mathrm{test}}}}

\DeclareMathAlphabet{\mathsfit}{\encodingdefault}{\sfdefault}{m}{sl}
\SetMathAlphabet{\mathsfit}{bold}{\encodingdefault}{\sfdefault}{bx}{n}

